%% file: main.tex
\documentclass[letterpaper, 10 pt, conference]{ieeeconf}
\IEEEoverridecommandlockouts 
\usepackage{graphicx} 
\usepackage{mathrsfs}
\usepackage{amsmath}
\usepackage{bm}
\usepackage{authblk}
\usepackage{amssymb}
\usepackage{booktabs}
\usepackage{hyperref}
\usepackage{amsfonts}
\DeclareGraphicsExtensions{.pdf,.png,.jpg}

\usepackage{xcolor}
\usepackage{standalone}
\usepackage{subcaption}

\usepackage{tikz}

\usepackage{amsthm}

\newtheorem{assumption}{Assumption}

\usetikzlibrary{angles,quotes}
\usepackage[backend=biber,style=ieee,natbib=true]{biblatex}
\title{Model Predictive Control for Magnetically-Actuated Cellbots \thanks{This work was partially supported by the NSF under grant GCR 9500313838.}}
\date{}
\author{
    Mehdi Kermanshah$^1$,
    Logan E. Beaver$^2$,
    Max Sokolich$^3$, 
    Fatma Ceren Kirmizitas$^{3,4}$,\\
    Sambeeta Das$^3$,
    Roberto Tron$^1$,
    Ron Weiss$^5$,
    Calin Belta$^6$
\thanks{$^1$Department of Mechanical Engineering,
    Boston University, Boston, MA, USA. Email:\{mker,tron\}@bu.edu}.
\thanks{$^2$Department of Mechanical and Aerospace Engineering, Old Dominion University, Norfolk, VA, USA. Email:lbeaver@odu.edu}
\thanks{$^3$Department of Mechanical Engineering, University of Delaware, Newark, DE, USA. Email:\{samdas,sokolich\}@udel.edu}
\thanks{$^4$Department of Animal and Food Sciences, University of Delaware, Newark, DE, USA.  Email:ceren@udel.edu}
\thanks{$^5$Department of Biological Engineering, MIT, Boston, MA, USA. Email: rweiss@mit.edu}
\thanks{$^6$Department of Electrical and Computer Engineering and Department of Computer Science, University of Maryland, College Park, MD, USA. Email:calin@umd.edu.}

}

\begin{document}
    
\maketitle

\begin{abstract}
 This paper presents a control framework for magnetically actuated cellbots, which combines Model Predictive Control (MPC) with Gaussian Processes (GPs) as a disturbance estimator for precise trajectory tracking. To address the challenges posed by unmodeled dynamics, we integrate data-driven modeling with model-based control to accurately track desired trajectories using relatively small data. To the best of our knowledge, this is the first work to integrate data-driven modeling with model-based control for the magnetic actuation of cellbots. The GP effectively learns and predicts unmodeled disturbances, providing uncertainty bounds as well. We validate our method through experiments with cellbots, demonstrating improved trajectory tracking accuracy. 
\end{abstract}  

\section{Introduction}

The use of micron-scale robots ($\mu$bots) for medical applications is an active area of research. 
Recent work includes using $\mu$bots for drug delivery  \cite{sitti2015biomedical, troccaz2008development}, biopsy \cite{Barcena2009ApplicationsBiomedicine}, microsurgery \cite{Guo2007MechanismApplication}, and cellular manipulation \cite{Sakar2011WirelessMicrotransporters,Jager2000paper,kim2013fabrication,Steager2013AutomatedMicrorobots}. Magnetically-actuated $\mu$bots are particularly appealing for medical applications due to their biocompatibility and the current use of magnets in clinical settings \cite{salehizadeh2020three}. However, precise motion control of magnetically-actuated $\mu$bots in the presence of noise and disturbances, especially in cluttered environments with obstacles, such as biological systems, remains a significant challenge \cite{Yang2020}. This challenge intensifies when $\mu$bots are embedded within living cells (cellbots), which is the focus of this study. 

Several model-based approaches have been developed for trajectory tracking using $\mu$bots \cite{jiang2022control}. Examples include linear and non-linear control approaches, including disturbance observers such as high-gain and extended state observers \cite{yang2018model}. Robust control techniques have also been employed to handle uncertainties in the system dynamics \cite{marino2013robust, liu2023adaptive}. Model Predictive Control (MPC) has been implemented to accurately track desired paths while minimizing control effort and ensuring adherence to output constraints \cite{jiang2022control, yang2019automated, yang2021autonomous, pieters2016model}. However, the performance of MPC depends heavily on the accuracy of the system model and the estimation of disturbances acting on the $\mu$bot. Traditional disturbance observers have been combined with MPC \cite{yang2021autonomous, belharet2010mri} to improve model accuracy. These approaches are limited in their ability to capture more complex dynamics. Alternatively, data-driven methods such as radial basis function neural networks \cite{liu2023adaptive} have been used to model disturbances and unknown dynamics. In our previous study \cite{beaver2022first}, we demonstratef that Gaussian Process (GP) is a more suitable choice as a disturbance estimator.


The works mentioned about apply solely to $\mu$bots, and are not suited to deal with the more complex dynamics of cellbots.
Related studies exploring $\mu$bots embedded within living cells \cite{chen2024cell, chen2022recent} utilize different robot designs \cite{lee2018fabrication, fan2020ferrofluid} and actuation mechanisms \cite{gwisai2022magnetic}. An earlier study used a rolling magnetic robot within a cell. However, it was limited to simple scenarios where the cellbot reached a target location in a straight line at a constant speed \cite{feng2020novel}.


In this work, we propose a framework for precise trajectory tracking of magnetically-actuated cellbots. We extend our previous work on data-driven $\mu$bot modeling \cite{Beaver2023LearningBots} by augmenting a MPC scheme and we estimate dynamic disturbances with GPs. A limitation of our previous approach is the initial data collection phase, which poses a collision risk in highly cluttered environments due to model inaccuracies. In this paper, by incorporating disturbance estimation into the MPC framework, we combine the advantages of both data-driven and model-based methods for accurate trajectory tracking. GPs are particularly well-suited for this task due to their ability to learn quickly from small datasets, allowing for individualized retraining of models for each cellbot. Additionally, the confidence intervals provided by GPs can be further utilized to ensure system safety under dynamic uncertainty. We conduct experiments with cellbots to demonstrate the effectiveness of our approach in real-world biological scenarios.
To the best of our knowledge, this is the first approach utilizing MPC enhanced with data-driven modeling for cellbot control. Our controller shows improved performance compared to previous state-of-the-art methods.



The remainder of this article is organized as follows. In Sec. \ref{sec:GP}, we provide a brief overview of Gaussian Processes. The problem formulation and an outline of our approach are given in Sec. \ref{sec:problem Formulation}. 
Our technical solution is presented in Sec. \ref{sec:solution}. 
Sec. \ref{sec:Experiment} shows the experimental results with a cellbot. Finally, in Sec. \ref{sec:conclusion}, we conclude with discussions and potential directions for future work.

\section{Mathematical Preliminaries}
\label{sec:GP}

In this section, we provide some background on GPs. Following \cite{seeger2004gaussian},
a GP is a stochastic process that generalizes the concept of a Gaussian distribution to a function space.
Formally, a GP is an infinite collection of random variables, any finite number of which have a joint Gaussian distribution.
A GP $y(\bm{x})$ is completely described by its \textit{mean} $m(\bm{x})$ and \textit{covariance} $k(\bm{x}, \bm{x}')$ functions.
We denote a GP using the compact form:
\begin{equation}
    y(\bm{x}) \sim \mathcal{GP}\big(m(\bm{x}), k(\bm{x}, \bm{x}')\big).
\end{equation}
In this paper, the feature vector $\bm{x}$ corresponds to the state and control variables of the cellbot (described in Section \ref{sec:problem Formulation}),
while the function $y(\bm{x})$ captures the unmodeled dynamics and stochastic disturbances acting on the cellbot.
We consider a zero-mean ($m(\bm{x})=0$) GP with noiseless observations, which is not restrictive in general \cite{seeger2004gaussian}.

The shape of the GP is completely defined by the covariance function $k(\bm{x}, \bm{x}')$, which measures the distance between two features. 
Furthermore, we can construct a covariance matrix $K$, where each entry $k_{ij} = k(\bm{x}_i, \bm{x}_j)$ and $i, j$ index the observations. 
Finally, by construction, we can predict the probability distribution of any finite number of disturbances $\bm{y}_*$ from predicted values of $\bm{x}_*$ given finitely many noiseless observations of $\bm{x}$ and $\bm{y}(\bm{x})$, i.e.,
\begin{equation}
    \begin{bmatrix}
        \bm{y} \\
        \bm{y}_*
    \end{bmatrix}
    \sim
    \mathcal{N}
    \left(
    \bm{0},
    \begin{bmatrix}
    K(\bm{x}, \bm{x}) & K(\bm{x}, \bm{x}_*) \\
    K(\bm{x}_*, \bm{x}) & K(\bm{x}_*, \bm{x}_*)
    \end{bmatrix}
    \right).
\end{equation}
The distribution of $\bm{y}_*$ conditioned on our previous observations and $\bm{x}_*$ is
\begin{equation}
\begin{aligned}
   & \mathbb{P}\left(\bm{y}_* \big| \bm{x}_*,\bm{x},\bm{y}\right) =  \mathcal{N}\Big(
        K(\bm{x}_*,\bm{x}) K(\bm{x},\bm{x})^{-1} \bm{y}, \\
        &\quad K(\bm{x}_*,\bm{x}_*) - K(\bm{x}_*,\bm{x})K(\bm{x},\bm{x})^{-1}K(\bm{x},\bm{x}_*)
    \Big),
\end{aligned}
\end{equation}
which is a probability distribution over the unmodeled dynamics and disturbances $\bm{y}_*$ at a predicted value of $\bm{x}_*$ based on historical observations of $\bm{y}, \bm{x}$.


 
\section{PROBLEM FORMULATION AND APPROACH}
\label{sec:problem Formulation}

\begin{figure}[h!]
    \centering
    \includestandalone[width=0.45\textwidth]{pipeline_fig} 
    \caption{Overview of the Control Framework: Initially, experimental data is collected offline to estimate $\hat{a}_0$ and train the GP for $\hat{\bm{D}}$. These are then used in an MPC framework to model the system dynamics. 
    }
    \label{fig:pipeline}
\end{figure}

\begin{figure}[htbp]
    \centering
     \begin{subfigure}[b]{0.23\textwidth}
    \centering
    \includegraphics[width=\textwidth]{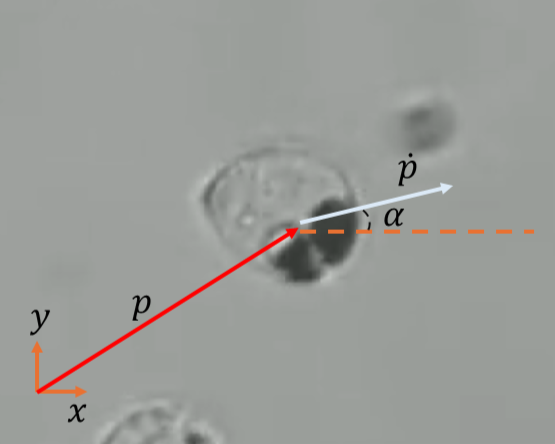}
    \caption{}
    \label{fig:cellbots}
  \end{subfigure}
  \hfill
  \begin{subfigure}[b]{0.23\textwidth}
    \centering
    \includegraphics[width=\textwidth]{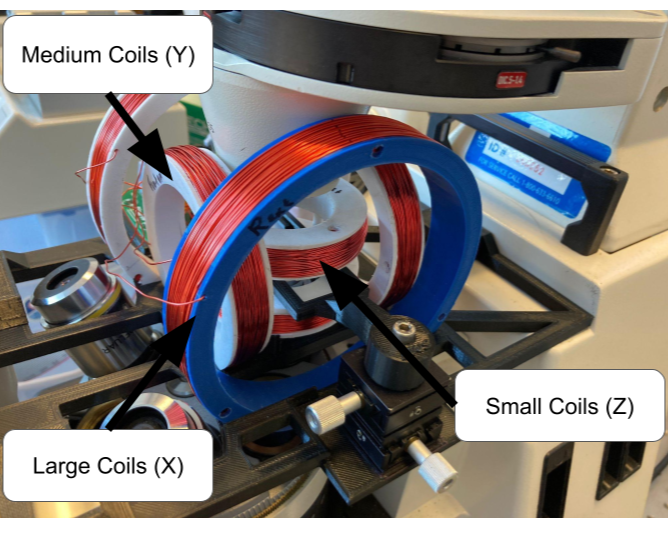}
    \caption{}
    \label{fig:experimental-setup}
  \end{subfigure}
    \caption{(a) Microscopy photo of a cellbot, showing the $\mu$bots (black disks) inside the cell; (b)The experimental setup used for cellbot control.}
\end{figure}

This work focuses on steering a cellbot to track desired trajectories by manipulating a magnetic field using an MPC framework. The experimental setup is shown in Fig. 
\ref{fig:experimental-setup} (see Sec. \ref{fig:experimental-setup} for details). 
An example of a cellbot, where $\mu$bots are injected into a cell, is shown in Fig. \ref{fig:cellbots}. 
We make the following assumption:

\begin{assumption}
\label{assumption:cell_coupled}
    The cells are mechanically coupled with the inserted $\mu$bots. The heading of a cellbot is determined by heading angle of the injected $\mu$bots.
\end{assumption}

Under the above assumption, a cellbot behaves as a rigid disk-shaped body, with its heading angle instantaneously aligning with that of the embedded $\mu$bots. 

Based on our experimental results and earlier studies 
\cite{beaver2022first}, we also make the following assumption: 

\begin{assumption} \label{assumption:2}
Cellbots align instantaneously and perfectly with the magnetic field.
\end{assumption}

Based on the above assumptions, we model the cellbot as a unicle, in which the heading can be instantaneously controlled, and which is subject to a generalized disturbance \cite{yang2019automated}:
\begin{equation}
\label{eq:dyn}
    \dot{\bm{p}} = a_0f(t) \begin{bmatrix}
\sin(\alpha(t)) \\ \cos(\alpha(t))
\end{bmatrix} + \bm{D}(t),
\end{equation}
where $\bm{p} \in \mathbb{R}^2, a_0 \in \mathbb{R}_{\geq 0} , f \in \mathbb{R}_{\geq 0}$ and $0 \leq \alpha < 2 \pi $ denote the position of the center of the cellbot, effective radius of rotation, frequency of the rotating magnetic field, and cellbot heading angle, respectively. The term $\bm{D} \in \mathbb{R}^2$ models the disturbance, capturing unmodeled dynamics such as Brownian motion, which is prevalent at the microscale.
Essentially, $a_0 f$ determines the forward speed and $\alpha$ sets the direction.  
In \eqref{eq:dyn},  the control inputs are $f$ and $\alpha$. 

The desired trajectory is specified as a finite set of waypoints $\bm{r}_{0:T-1} = \{\bm{r}_0, \bm{r}_1, \ldots, \bm{r}_{T-1}\}$, where $\bm{r}_t$  denotes the waypoint at (discrete) time $t$, $t=0,\ldots T-1$. 
Our goal is to find $f$ and $\alpha$ for system \eqref{eq:dyn} such that the following cost is minimized while the control bounds are satisfied:

\begin{equation}
\label{eq:lossfucntion}
\begin{aligned}
      &\min_{f_{0:T-1},\alpha_{0:T-1}, \bm{p}_{0:T-1}} \sum_{t=0}^{T-1} \left( \gamma^t(\bm{p}_t - \bm{r}_t)^T Q (\bm{p}_t - \bm{r}_t) + R f_t^2 \right)\\
      &\text{subject to} \quad 0 \leq f_t \leq f_\text{max}
\end{aligned}
\end{equation}

In the cost function, \(Q\) and \(R\) are weight matrices that balance tracking accuracy with control effort, and \(f_{\text{max}}\) represents the frequency limit. We introduce a discount factor \(\gamma\) to mitigate the influence of future predictions on current actions,since these predictions are less accurate due to the system's stochasticity. In \eqref{eq:lossfucntion}, we penalize \(f\) to ensure trajectory tracking at lower frequencies to avoid exciting disturbances at higher frequencies.

Our proposed approach is illustrated in Fig. \ref{fig:pipeline}. First, during the initial learning phase, we collect data consisting of the pairs $(\dot{p}, f, \alpha)$. Using this data, we estimate $a_0$ in the dynamics model \eqref{eq:dyn} through linear regression. Next, we utilize the GP to learn the error between the unicycle model and the collected data. MPC is then adopted to derive the control action, minimizing the cost function \eqref{eq:lossfucntion} over the time horizon $T_h$ using the GP-enhanced model.


\begin{figure}[h!]
  \centering
  \begin{subfigure}[b]{0.48\textwidth}
    \centering
    \includegraphics[trim=40 50 40 70,clip,width=\textwidth]{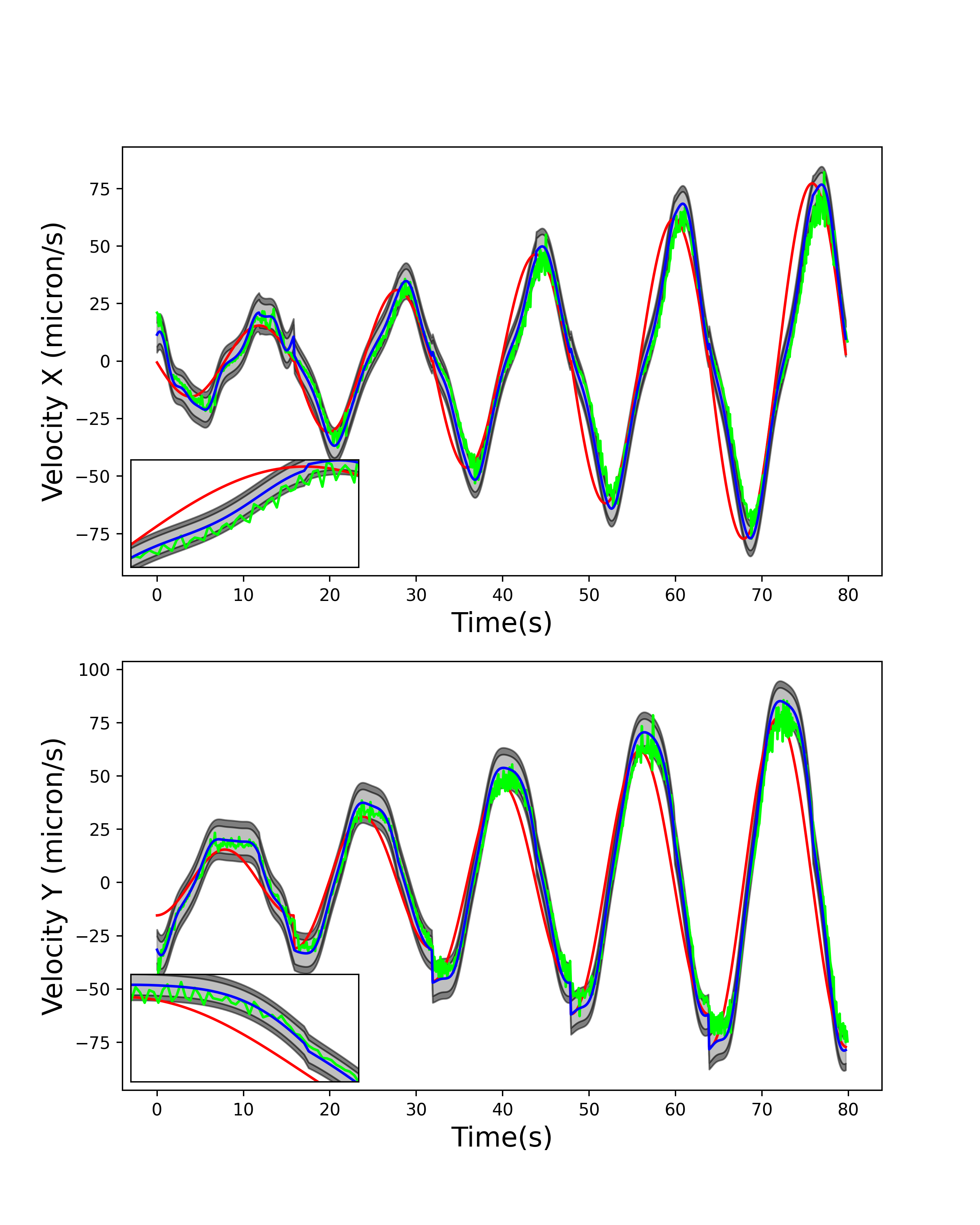}
  \end{subfigure}
  \caption{Comparison of the learned model, actual data, and the linear model with GP disturbance. The top and bottom graphs correspond to the x and y directions, respectively.  The inset plots zoom in on the time range between 40 and 45 seconds. The blue, red and green curves represent the learned model, linear model without GP and the actual velocity($\mu m/s$), respectively.  The light and dark gray bands indicate two standard deviations (65\%) of uncertainty and three standard deviations (95\%).
  }
  \label{fig:traindata}
\end{figure}

\begin{figure*}
  \centering
  \begin{subfigure}[b]{0.48\textwidth}
    \centering
    \includegraphics[trim=6 60 50 85,clip,width=\textwidth]{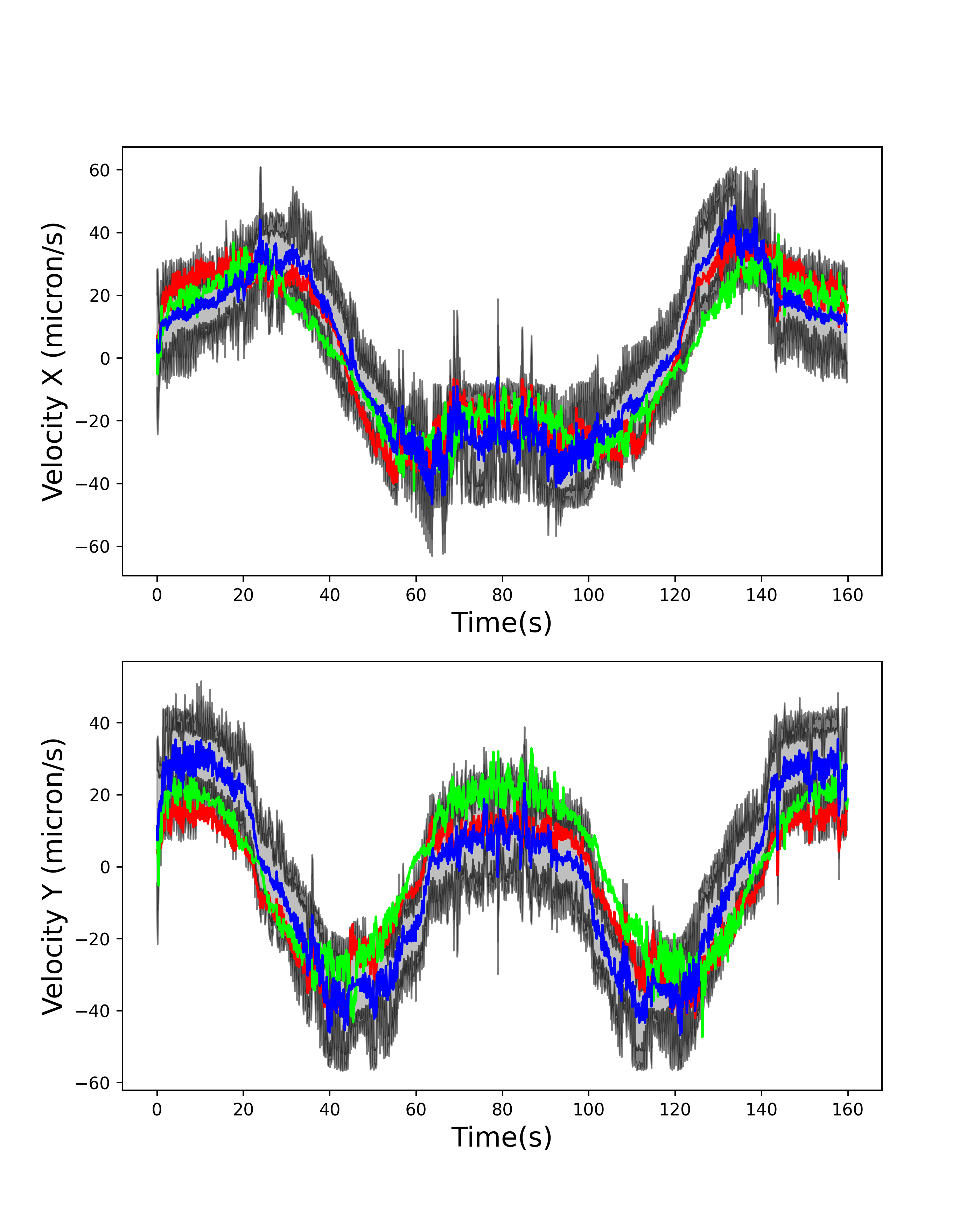}
    \caption{}
    \label{fig:firsttraining_vel}
  \end{subfigure}
  \hfill
  \begin{subfigure}[b]{0.48\textwidth}
    \centering
    \includegraphics[trim=5 60 50 85,clip,width=\textwidth]{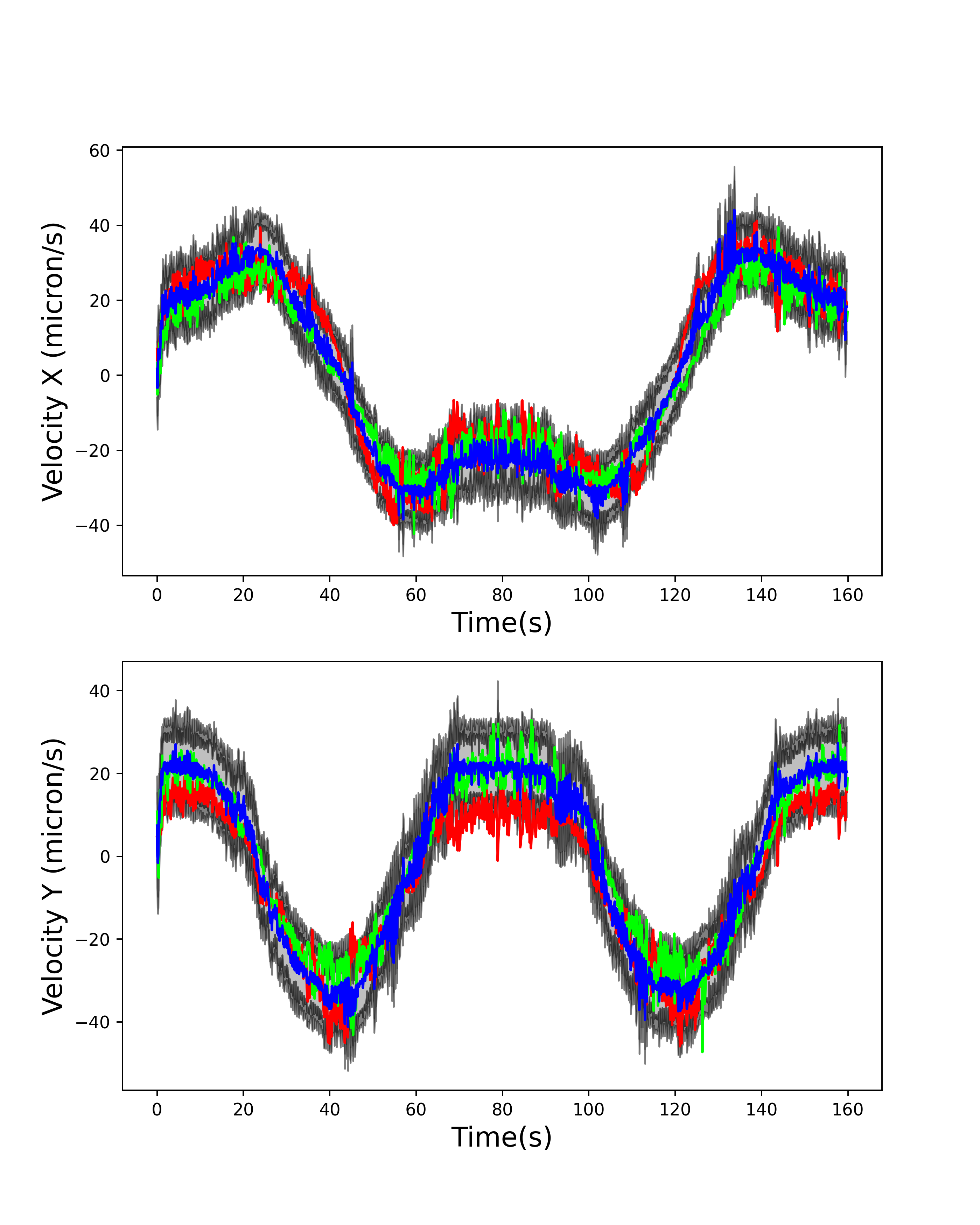}
    \caption{}
    \label{fig:it_training_vel}
  \end{subfigure}
  \hfill
  \caption{Model predictions after the circular (a) and iterative training (b), respectively. 
  The top and bottom graphs correspond to the x and y directions, respectively. Iterative training noticeably enhances the model's accuracy. We used the same color coding as in Fig \ref{fig:traindata}. 
  }
  \label{fig:mpcexperiment}
\end{figure*}

\section{Model Predictive Control with GP Disturbance Estimator}
\label{sec:solution}

 The following assumption is made in order to simplify learning the disturbance from the data:

\begin{assumption} \label{assumption:1}
The environmental disturbance $\bm{D}$ and any error in our model of the true dynamics \eqref{eq:dyn} are isotropic, i.e., they are not a function of the cellbot's position $p$.
\end{assumption}


 We model the disturbance as a function of frequency and heading angle (i.e., $\bm{D}(\alpha, f)$). 

\subsection{Disturbance Estimation with GP}

The system dynamics \eqref{eq:dyn} is nonlinear in $\alpha$, which makes the above optimization problem nonconvex. To reduce computational complexity, we transform the system from polar to Cartesian coordinates using $\bm{u} = [ u_x, u_y]^T = [f\cos(\alpha), f\sin(\alpha)]^T$:
\begin{equation}
\label{eq:lin_dyn}
    \dot{\bm{p}} = a_0 \begin{bmatrix}
u_x \\ u_y
\end{bmatrix} + \bm{D}(\alpha, f) = a_0 \bm{u} + \bm{D}(\alpha, f).
\end{equation}
Note that in \eqref{eq:lin_dyn}, $\bm{u}$ can be mapped to $f$ and $\alpha$ as:
\begin{align}
    f = \sqrt{\bm{u}^T \bm{u}}, && \alpha =  \text{atan}2(u_y/u_x).
\end{align}

According to Assumption \ref{assumption:2}, the coordinate transformation does not introduce any kinematic constraints (like the minimum turning radius) since the electrical system for varying the magnetic field responds instantaneously.

In this section, we describe our approach for estimating \(\bm{D}\) and \(a_0\). We train the GP during an initial learning phase, where the cellbot is given a sequence of control inputs, either from a human operator or a pre-defined path. The learning phase occurs in the same environment and with the same robot as the experiment; thus, the training and testing environments are consistent. 
Position data ($\mathcal{P} = \{[p_x(t_k), p_y(t_k)]^T\}$) and control action data ($X = \{(\alpha(t_k), f(t_k))\}$) are recorded at discrete intervals.  We derive the cellbot velocity ($\mathcal{V} = \{[v_x(t_k), v_y(t_k)]^T\}$) by numerical differentiation and low-pass filtering of $\mathcal{P}$. Based on the dynamics, we do a linear regression separately for the $x$ and $y$ axes:
\begin{equation}
\begin{aligned}
\label{eq:linreg}
    v_x  &= \hat{a}_{0,x}u_x + \hat{D}_{c,x}, \\
    v_y  &= \hat{a}_{0,y}u_y + \hat{D}_{c,y}.
\end{aligned}
\end{equation}
 Note that in \eqref{eq:linreg}, we assumed that $\hat{\bm{D}}$ is constant and  $\bm{D_c} = [ D_{c,x}, D_{c,y}]^T$ can be interpreted as the mean disturbance value, as estimated by the linear regression model. We use linear regression to estimate $a_0$ by $\hat{a}_0$ and utilize the GP to estimate the model mismatch. Based on \eqref{eq:linreg}, we estimate $a_0$ by
 \begin{equation}
     \hat{a}_0 =\frac{1}{\sqrt{2}} \sqrt{\hat{a}_{0,x}^2+\hat{a}_{0,y}^2}.
 \end{equation}
Based on this estimation, we can rewrite the dynamics as follows:
\begin{equation}
\label{eq:dyn_ve}
    \dot{\bm{p}} = \hat{a}_0\bm{u} + \bm{v_e}.
\end{equation}

We estimate $\bm{v_e}$, the model mismatch, from data using a GP. The model mismatch \(\bm{v_e}\) consists of two components: the first is the model uncertainty $D$ in \eqref{eq:dyn}, and the second is the error in estimating \(a_0\). Consequently, compared to the original dynamics, the model mismatch is expressed as $\bm{v_e} = (a_0 - \hat{a}_0)\bm{u} + \bm{D}$.
 The GP essentially maps $X$ to the observed error velocities denoted by $\mathcal{Y} = \{(\bm{v}(t_k) - \hat{a}_0 \bm{u}(t_k))\}$. 
We implement this approach similarly to \cite{beaver2022first}, utilizing a linear combination of the Radial Basis Function (RBF) kernel and a constant-scaled white noise kernel, which effectively captures the velocity error. Our kernel function can be formally expressed as:

\begin{equation}
K(\bm{x}, \bm{x}^\prime) = C\left(\exp\left(-\frac{||\bm{x} - \bm{x}^\prime||^2}{2\sigma}\right) + \eta\right), \label{eq:kernel}
\end{equation}
where $\sigma$ and $C$ are length scale and constant scale hyperparameters, and $\eta$ is drawn from a normal distribution with zero mean and variance as another hyperparameter.

\subsection{Model Predictive Control}

To generate trajectories, we discretize the dynamics using Euler's method:
\begin{equation}
    \bm{p}_{t+1} = \bm{p}_{t} + \hat{a}_0 \Delta t \bm{u}_t + \hat{\bm{v}}_{\bm{e}} \Delta t
\end{equation}
where $\bm{p}_t$, $\bm{u}_t$, $\hat{\bm{v}}_{\bm{e}}$, and $\Delta t$ are the position, control, estimation of the model mismatch by the GP, and time step, respectively.


By rewriting \eqref{eq:lossfucntion} for the discretized system, the control action at each time step is derived from the following optimization problem:
\begin{equation}
\label{eq:MPCopt}
\begin{aligned}
    \min_{\bm{u}_{0:T_h-1}, \bm{p}_{0:T_h-1}} &\sum_{t=0}^{T_h-1} \left[ \gamma^t (\bm{p}_t - \bm{r}_t)^T Q (\bm{p}_t - \bm{r}_t) + \bm{u}_t^T R \bm{u}_t \right] \\
    &\text{subject to} \\&
     \bm{p}_t =  \bm{p}_{t-1} + \hat{a}_0 \Delta t \bm{u}_t + \hat{\bm{v}}_{\bm{e}} \Delta t,  \quad t \in [0, T_h-1] \\
    & \bm{u}_{\text{min}} \leq \bm{u}_t \leq \bm{u}_{\text{max}},  \quad t \in [0, T_h-1].
\end{aligned}
\end{equation}
\begin{figure*}
  \centering
  \begin{subfigure}[b]{0.48\textwidth}
    \centering
    \includegraphics[width=\textwidth]{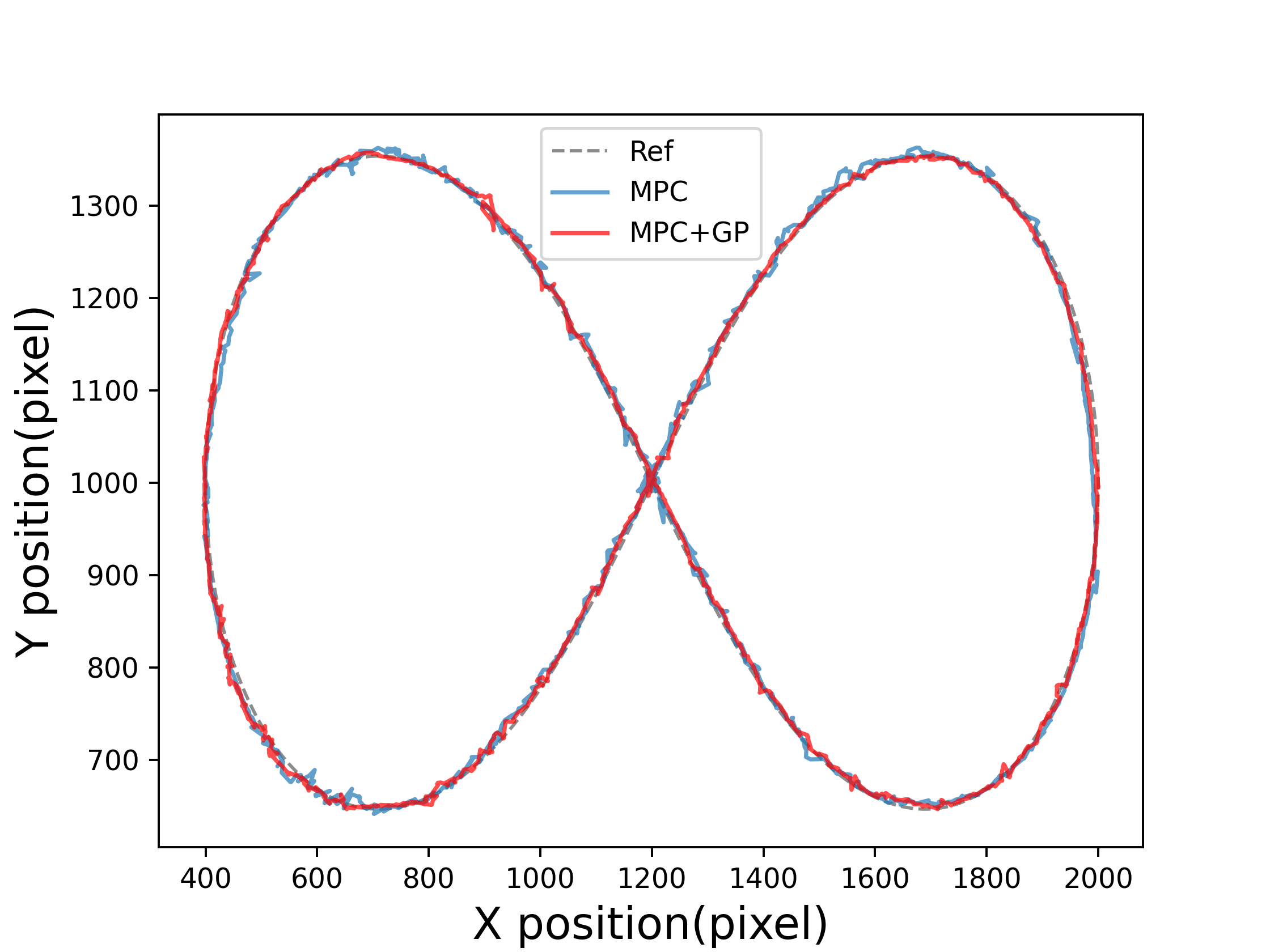}
    \caption{}
    \label{fig:comparewithMPC}
  \end{subfigure}
  \hfill
  \begin{subfigure}[b]{0.48\textwidth}
    \centering
    \includegraphics[width=\textwidth]{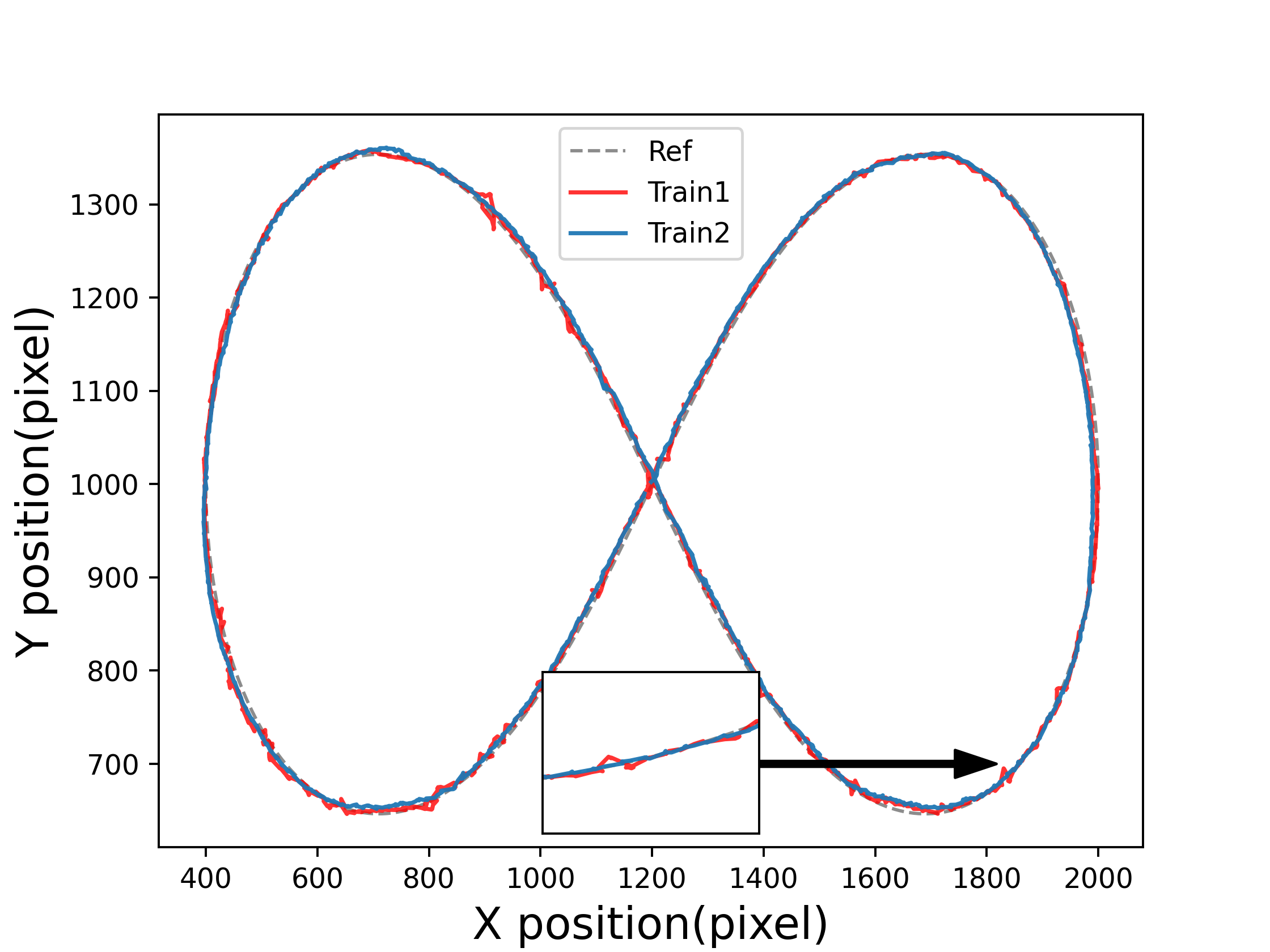}
    \caption{}
    \label{fig:it_traj}
  \end{subfigure}
  \caption{ 
   Trajectories after circular (a) and iterative (b) training. In (b), the inset zoom in on the lower left shows a closer look at the differences in the trajectories. 
   }
 
\end{figure*}

We estimate the disturbance using GP based on the current values of \(f\) and \(\alpha\), and keep the estimated disturbance $\hat{\bm{v}}_{\bm{e}}$ constant throughout the predictive horizon. This approach simplifies the optimization into a quadratic problem. However, this simplification reduces the accuracy of long-term prediction. The second constraint captures the limit on $f$. We use this to define the range within which the GP is trained to estimate disturbances.

\section{Results}
\label{sec:Experiment}




\subsection{Experimental Setup}
\label{sec:exp-setup}

We embed rolling microrobots, as in \cite{Beaver2023LearningBots}, within the cells described below. The cell culture components are purchased from GIBCO. Chinese Hamster Ovary (CHO) cells are cultured in Dulbecco's Modified Essential Medium-F12 (DMEM-F12) with 10\% Fetal Bovine Serum (FBS), 1\% penicillin/streptomycin, 5\% CO\(_2\), and incubated at 37°C. After washing with Dulbecco's phosphate-buffered saline (DPBS) and trypsinizing to detach, cells from the third passage are used. CHO cells are seeded into a 35 mm petri dish at a density of \(1 \times 10^5\) cells/dish and incubated for 24 hours. Microrobots are dispersed in the culture media and added to the petri dish. After incubation, cells are washed, trypsinized, and centrifuged to collect the cellbots. To prevent the cellbots from sticking to the surface, they are suspended in a 0.1\% sodium dodecyl sulfate solution instead of DI water. Although this increases viscosity and reduces rolling speed, it significantly reduces surface adhesion.

Six Helmholtz coils, arranged in parallel pairs (Fig. \ref{fig:experimental-setup}), generate the magnetic fields needed to actuate the cellbot. The coils are mounted on a Zeiss Axiovert 100 inverted microscope and powered by an Arduino Mega microcontroller connected to a Desktop computer. A Python-based control program processes images from a FLIR BFS-U3-28S5M-C USB camera via OpenCV to extract position and velocity data for the cellbot.

\subsection{Experiment}

\label{sec:experiments}

First, we collect data during an initial training phase to learn the disturbance experienced by the robot in the environment. We collect experimental data within the frequency range \(0 \leq f \leq 5 \, \text{Hz}\) with increments of 1 Hz, and angular range \(0 \leq \alpha \leq 2\pi\) with increments of 1 degree. We call this circular training dataset since when the frequency is fixed, the robot moves in a circular path, and the radius of the circle changes with the frequency. The videos of the experiments, along with supplementary material, is available online\footnote{\url{https://sites.google.com/bu.edu/mpcgp}}.
Note that we perform training and testing in the same environment. Thus, we do not consider issues that may arise from policy transfer or environmental inconsistency.

We train two separate GPs to estimate the model mismatch in the x and y directions, both using the kernel function defined in \eqref{eq:kernel}. To evaluate the effectiveness of our approach, we compared the GP-enhanced model against the actual collected data and a baseline linear model without the GP disturbance estimator, as shown in Fig. \ref{fig:traindata}. The GP models effectively captured disturbance behavior, leading to a significant enhancement in the accuracy of the linear model.

In our experiments, we set \(Q = I\), \(R = 0.01 \times I\), \(\Delta t = 0.1\), and a prediction horizon \(T_h = 4\), where $I$ denotes the identity matrix. The desired trajectory for this case study is a Lemniscate of Bernoulli path, as depicted in Fig. \ref{fig:comparewithMPC}. The desired and actual trajectories are represented by gray and red curves, respectively. Our approach successfully tracks the desired trajectory with high accuracy. In the same figure, we compare our method with a naive MPC approach that does not utilize the GP disturbance estimator. As shown, our approach tracks the desired trajectory more smoothly and more accurately. It achieves a 14\% reduction in the 2-norm mean tracking error and a 28\% reduction in the maximum deviation(2-norm error) from the desired trajectory.



We also plot our model's predictions in Fig \ref{fig:firsttraining_vel}. In some areas, the GP disturbance prediction closely aligns with the linear model. This indicates that the MPC controller drives the system into $\alpha$ and $f$ pairs were not observed in the training data due to sampling only integer frequencies in the circular training dataset.

\subsection{Iterative Learning}
To improve tracking accuracy, we collect additional data (pairs of ($\alpha, f, \dot{p}$)) through an iterative process \cite{wagener2019online}. We use MPC with a GP disturbance estimator to track the desired trajectory, which in our case study is the Lemniscate of Bernoulli. All data points not observed in the initial circular dataset are then augmented in the new training dataset. This process can be repeated to explore the control space more effectively. We found that performing this process just once significantly improves both our model's performance and tracking accuracy, as shown in Fig. \ref{fig:it_training_vel}.

\begin{figure}[htbp]
  \begin{subfigure}[b]{0.48\textwidth}
    \centering
    \includegraphics[width=\textwidth]{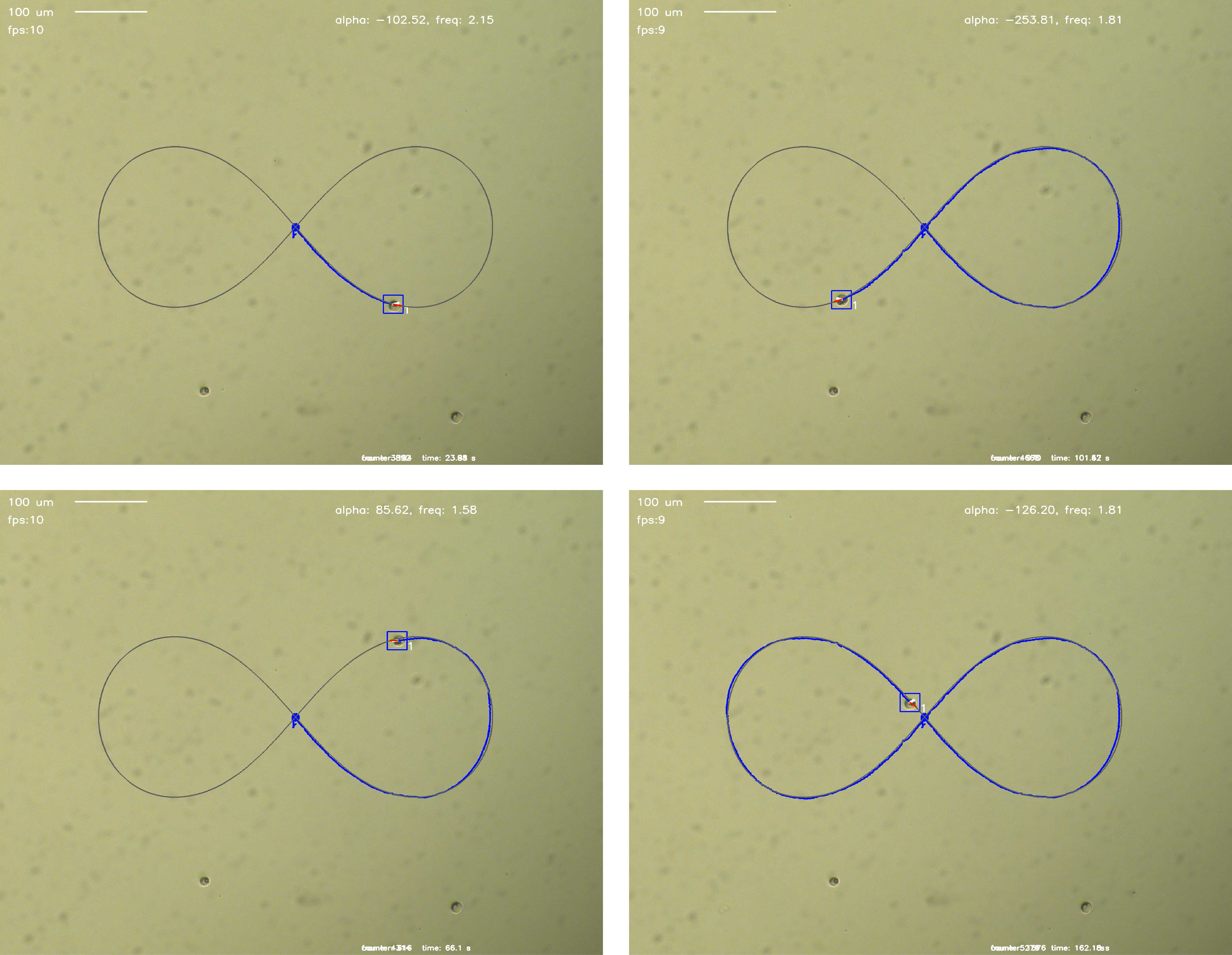}
  \end{subfigure}
  \caption{Snapshots showing the progress of the cellbot tracking a Lemniscate of Bernoulli trajectory experiment. The square highlights the cellbot suspended in the aqueous solution. The history of the cellbot's movement is overlaid. The blue and gray curves represent the actual and desired trajectories, respectively. 
  }
  \label{fig:cy}
\end{figure}

In Fig. \ref{fig:it_traj}, the trajectories after the first and iterative training are shown in red and blue, respectively. The snapshots of the real system is shown in Fig. \ref{fig:cy}. After iterative training, our approach tracks the desired trajectory more accurately, with reduced deviation.
To evaluate the performance of the GP, we compare the Mean Absolute Error (MAE) of the learned model with that of the linear model, as shown in Table~\ref{tab:MAE}. After iterative training, the GP reduces the modeling error by approximately 60\%. Additionally, we present the output of our learned model with confidence intervals alongside the actual velocity in Figures~\ref{fig:firsttraining_vel} and~\ref{fig:it_training_vel}, corresponding to the first training and after iterative training, respectively.  After the final training, the learned model closely predicts the observed velocity. The actual velocity is within one standard deviation 56\% of the time and within two standard deviations 86\% of the time. Based on this, we can conclude that the Gaussian distribution is a close approximation for $D$, but it tends to under-approximate (e.g., 86\% vs. 96\% for $\pm 2$ standard deviations).
 Although this confidence bond has not yet been integrated into our framework, it offers a potential avenue to be combined in robust MPC frameworks, such as robust \cite{bemporad2007robust} or tube-based \cite{limon2010robust} MPC.

\begin{table}[htbp]
\centering
\caption{Mean absolute errors for the cellbot experiments and comparison of improvements relative to the baseline (MPC with GP disturbance estimator). The numbers in parentheses indicate the percentage improvement over the baseline.}
\label{tab:MAE}
\begin{tabular}{@{}lccc@{}}
\toprule
 & \textbf{MPC} (baseline) & \textbf{Circular Training} & \textbf{Iterative Training} \\ \midrule
\text{$x$ axis}   & 9.64 $\mu$m/s & 7.93 $\mu$m/s  (18$\%$) & 3.84 $\mu$m/s (60$\%$) \\ 
\text{$y$ axis}   & 10.18 $\mu$m/s & 10.07 $\mu$m/s (1$\%$) & 3.93 $\mu$m/s (61$\%$) \\ \bottomrule
\end{tabular}
\end{table}


\section{Conclusion}
\label{sec:conclusion}
This paper presents a control framework for cellbots that integrates MPC with Gaussian Processes for disturbance estimation. Our approach addresses the challenges of controlling a cellbot
in the presence of disturbances. By leveraging GP, we obtain accurate disturbance estimates with a relatively small amount of data. The effectiveness of GP allows us to retrain the model for different cellbots without requiring significant computing power. This approach combines the strengths of data-driven modeling with the robustness of model-based control, offering a practical solution for the precise and reliable control of microrobots. Our experimental results demonstrate that the proposed framework effectively mitigates disturbances and improves trajectory tracking accuracy.

Future work will involve using the GP-provided confidence intervals in planning algorithms with tube-based MPC to enhance robot safety. Additionally, we plan to explore online training, allowing the cellbot to update the GP model with new data points as it navigates its environment.
\printbibliography

\end{document}

%% file: pipeline_fig.tex
\begin{tikzpicture}
    \tikzstyle{block} = [rectangle, draw, text width=6em, text centered, rounded corners, minimum height=4em]
    
    \fill[gray!20] (-1,-0.5) rectangle (9,2);
    
    \node at (4,1.8) {Initial learning phase};
    
    \node [block] (data) at (1,0.5) {Experiment Data Collection};
    \node [block] (estimate) at (7,0.5) {Estimate \(a_0\), Train GP};
    
    \node [block] (mpc) at (4,-2) {MPC Controller};
    
    \draw [->] (data) -- (estimate);
    \draw [->] (4,0) -- (mpc);
    
    \draw [->] (1,-2) -- (mpc);
    \draw [->] (mpc) -- (7,-2);
      
    \node [above] at (1.2,-2) {Desired Trajectory};
    \node [above] at (1.8,-2.5) {\(r_{0:T_h-1}\)};
    \node [above] at (6.1,-2) {Control};
    \node [above] at (6.2,-2.5) { \(u_t\)};
\end{tikzpicture}

%% file: references.bib
@article{seeger2004gaussian,
  title={Gaussian processes for machine learning},
  author={Seeger, Matthias},
  journal={International journal of neural systems},
  volume={14},
  number={02},
  pages={69--106},
  year={2004},
  publisher={World Scientific}
}

@article{Beaver2023LearningBots,
  author={Beaver, Logan E. and Sokolich, Max and Alsalehi, Suhail and Weiss, Ron and Das, Sambeeta and Belta, Calin},
  journal={IEEE Robotics and Automation Letters}, 
  title={Learning a Tracking Controller for Rolling $\mu$bots}, 
  year={2024},
  volume={9},
  number={2},
  pages={1819-1826},
  doi={10.1109/LRA.2024.3350968}}

@inproceedings{beaver2022first,
  title={A first-order approach to model simultaneous control of multiple microrobots},
  author={Beaver, Logan E and Wu, Bingzhi and Das, Sambeeta and Malikopoulos, Andreas A},
  booktitle={2022 International Conference on Manipulation, Automation and Robotics at Small Scales (MARSS)},
  pages={1--7},
  year={2022},
  organization={IEEE}
}

@ARTICLE{Yang2020,
  author={Yang, Z and Zhang, L},
  title={Magnetic Actuation Systems for Miniature Robots: A Review},
  journal={Advanced Intelligent Systems}, 
  volume={2},
  year={2020}
}

@article{salehizadeh2020three,
  title={Three-dimensional independent control of multiple magnetic microrobots via inter-agent forces},
  author={Salehizadeh, Mohammad and Diller, Eric},
  journal={The International Journal of Robotics Research},
  volume={39},
  number={12},
  pages={1377--1396},
  year={2020},
  publisher={SAGE Publications Sage UK: London, England}
}

@article{sitti2015biomedical,
  title={Biomedical applications of untethered mobile milli/microrobots},
  author={Sitti, Metin and Ceylan, Hakan and Hu, Wenqi and Giltinan, Joshua and Turan, Mehmet and Yim, Sehyuk and Diller, Eric},
  journal={Proceedings of the IEEE},
  volume={103},
  number={2},
  pages={205--224},
  year={2015},
  publisher={IEEE}
}

@article{troccaz2008development,
  title={The development of medical microrobots: a review of progress},
  author={Troccaz, Jocelyne and Bogue, Robert},
  journal={Industrial Robot: An International Journal},
  year={2008},
  publisher={Emerald Group Publishing Limited}
}

@incollection{Barcena2009ApplicationsBiomedicine,
    title = {{Applications of magnetic nanoparticles in biomedicine}},
    year = {2009},
    author = {B{\'a}rcena, Carlos and Sra, Amandeep K. and Gao, Jinming},
    booktitle = {Nanoscale Magnetic Materials and Applications}
}

@inproceedings{Guo2007MechanismApplication,
    title = {{Mechanism and control of a novel type microrobot for biomedical application}},
    year = {2007},
    booktitle = {Proceedings 2007 IEEE International Conference on Robotics and Automation},
    author = {Guo, Shuxiang and Pan, Qinxue},
    pages = {187--192},
}

@inproceedings{Sakar2011WirelessMicrotransporters,
    title = {{Wireless manipulation of single cells using magnetic microtransporters}},
    year = {2011},
    booktitle = {2011 IEEE International Conference on Robotics and Automation},
    author = {Sakar, Mahmut Selman and Steager, Edward B and Cowley, Anthony and Kumar, Vijay and Pappas, George J},
    pages = {2668--2673}
}

@article{Jager2000paper,
    title = {{Microrobots for micrometer-size objects in aqueous media: potential tools for single-cell manipulation}},
    year = {2000},
    author = {Jager, Edwin W H and Ingan{\"{a}}s, Olle and Lundstr{\"{o}}m, Ingemar Science},
    number = {5475},
    pages = {2335--2338},
    volume = {288}
}

@article{kim2013fabrication,
  title={Fabrication and characterization of magnetic microrobots for three-dimensional cell culture and targeted transportation},
  author={Kim, Sangwon and Qiu, Famin and Kim, Samhwan and Ghanbari, Ali and Moon, Cheil and Zhang, Li and Nelson, Bradley J and Choi, Hongsoo},
  journal={Advanced Materials},
  volume={25},
  number={41},
  pages={5863--5868},
  year={2013},
  publisher={Wiley Online Library}
}

@article{Steager2013AutomatedMicrorobots,
    title = {{Automated biomanipulation of single cells using magnetic microrobots}},
    year = {2013},
    author = {Steager, Edward B and Selman Sakar, Mahmut and Magee, Ceridwen and Kennedy, Monroe and Cowley, Anthony and Kumar, Vijay},
    booktitle = {The International Journal of Robotics Research},
    number = {3},
    pages = {346--359},
    volume = {32}
}

@article{marino2013robust,
  title={Robust electromagnetic control of microrobots under force and localization uncertainties},
  author={Marino, Hamal and Bergeles, Christos and Nelson, Bradley J},
  journal={IEEE Transactions on Automation Science and Engineering},
  volume={11},
  number={1},
  pages={310--316},
  year={2013},
  publisher={IEEE}
}

@article{jiang2022control,
  title={Control and autonomy of microrobots: Recent progress and perspective},
  author={Jiang, Jialin and Yang, Zhengxin and Ferreira, Antoine and Zhang, Li},
  journal={Advanced Intelligent Systems},
  volume={4},
  number={5},
  pages={2100279},
  year={2022},
  publisher={Wiley Online Library}
}

@article{yang2018model,
  title={Model-free trajectory tracking control of two-particle magnetic microrobot},
  author={Yang, Lidong and Wang, Qianqian and Zhang, Li},
  journal={IEEE Transactions on Nanotechnology},
  volume={17},
  number={4},
  pages={697--700},
  year={2018},
  publisher={IEEE}
}

@article{yang2021autonomous,
  title={Autonomous navigation of magnetic microrobots in a large workspace using mobile-coil system},
  author={Yang, Zhengxin and Yang, Lidong and Zhang, Li},
  journal={IEEE/ASME Transactions on Mechatronics},
  volume={26},
  number={6},
  pages={3163--3174},
  year={2021},
  publisher={IEEE}
}

@article{belharet2010mri,
  title={MRI-based microrobotic system for the propulsion and navigation of ferromagnetic microcapsules},
  author={Belharet, Karim and Folio, David and Ferreira, Antoine},
  journal={Minimally Invasive Therapy \& Allied Technologies},
  volume={19},
  number={3},
  pages={157--169},
  year={2010},
  publisher={Taylor \& Francis}
}

@article{liu2023adaptive,
  title={Adaptive learning and sliding mode control for a magnetic microrobot precision tracking with uncertainties},
  author={Liu, Yueyue and Wang, Haoyu and Fan, Qigao},
  journal={IEEE Robotics and Automation Letters},
  year={2023},
  publisher={IEEE}
}

@article{wagener2019online,
  title={An online learning approach to model predictive control},
  author={Wagener, Nolan and Cheng, Ching-An and Sacks, Jacob and Boots, Byron},
  journal={arXiv preprint arXiv:1902.08967},
  year={2019}
}

@article{yang2019automated,
  title={Automated control of magnetic spore-based microrobot using fluorescence imaging for targeted delivery with cellular resolution},
  author={Yang, Lidong and Zhang, Yabin and Wang, Qianqian and Chan, Kai-Fung and Zhang, Li},
  journal={IEEE Transactions on Automation Science and Engineering},
  volume={17},
  number={1},
  pages={490--501},
  year={2019},
  publisher={IEEE}
}

@article{pieters2016model,
  title={Model predictive control of a magnetically guided rolling microrobot},
  author={Pieters, Roel and Lombriser, Sievi and Alvarez-Aguirre, Alejandro and Nelson, Bradley J},
  journal={IEEE Robotics and Automation Letters},
  volume={1},
  number={1},
  pages={455--460},
  year={2016},
  publisher={IEEE}
}

@article{limon2010robust,
  title={Robust tube-based MPC for tracking of constrained linear systems with additive disturbances},
  author={Lim{\'o}n, Daniel and Alvarado, Ignacio and Alamo, TEFC and Camacho, Eduardo F},
  journal={Journal of Process Control},
  volume={20},
  number={3},
  pages={248--260},
  year={2010},
  publisher={Elsevier}
}

@incollection{bemporad2007robust,
  title={Robust model predictive control: A survey},
  author={Bemporad, Alberto and Morari, Manfred},
  booktitle={Robustness in identification and control},
  pages={207--226},
  year={2007},
  publisher={Springer}
}

@article{chen2022recent,
  title={Recent advances in field-controlled micro--nano manipulations and micro--nano robots},
  author={Chen, Yuanyuan and Chen, Dixiao and Liang, Shuzhang and Dai, Yuguo and Bai, Xue and Song, Bin and Zhang, Deyuan and Chen, Huawei and Feng, Lin},
  journal={Advanced Intelligent Systems},
  volume={4},
  number={3},
  pages={2100116},
  year={2022},
  publisher={Wiley Online Library}
}

@article{gwisai2022magnetic,
  title={Magnetic torque--driven living microrobots for increased tumor infiltration},
  author={Gwisai, Tinotenda and Mirkhani, Nima and Christiansen, Michael G and Nguyen, Thuy Trinh and Ling, V and Schuerle, SJSR},
  journal={Science Robotics},
  volume={7},
  number={71},
  pages={eabo0665},
  year={2022},
  publisher={American Association for the Advancement of Science}
}

@article{chen2024cell,
  title={Cell-Based Micro/Nano-Robots for Biomedical Applications: A Review},
  author={Chen, Bo and Sun, Hongyan and Zhang, Jiaying and Xu, Junjie and Song, Zeyu and Zhan, Guangdong and Bai, Xue and Feng, Lin},
  journal={Small},
  volume={20},
  number={1},
  pages={2304607},
  year={2024},
  publisher={Wiley Online Library}
}

@article{lee2018fabrication,
  title={Fabrication and characterization of a magnetic drilling actuator for navigation in a three-dimensional phantom vascular network},
  author={Lee, Sunkey and Lee, Seungmin and Kim, Sangwon and Yoon, Chang-Hwan and Park, Hun-Jun and Kim, Jin-young and Choi, Hongsoo},
  journal={Scientific reports},
  volume={8},
  number={1},
  pages={3691},
  year={2018},
  publisher={Nature Publishing Group UK London}
}

@article{fan2020ferrofluid,
  title={Ferrofluid droplets as liquid microrobots with multiple deformabilities},
  author={Fan, Xinjian and Sun, Mengmeng and Sun, Lining and Xie, Hui},
  journal={Advanced Functional Materials},
  volume={30},
  number={24},
  pages={2000138},
  year={2020},
  publisher={Wiley Online Library}
}

@inproceedings{feng2020novel,
  title={A novel and controllable cell-based microrobot in real vascular network for target tumor therapy},
  author={Feng, Yanmin and Feng, Lin and Dai, Yuguo and Bai, Xue and Zhang, Chaonan and Chen, Yuanyuan and Arai, Fumihito},
  booktitle={2020 IEEE/RSJ International Conference on Intelligent Robots and Systems (IROS)},
  pages={2828--2833},
  year={2020},
  organization={IEEE}
}
